\documentclass{article}
\usepackage{spconf,amsmath,graphicx}
\usepackage{subcaption}
\usepackage{arydshln}
\usepackage{hyperref}
\usepackage{amsfonts}
\usepackage{multirow}
\usepackage{blindtext}
\usepackage{comment}

\newcommand{\quotes}[1]{``#1''}
\newcommand{\squeezeup}{\vspace{-0.7cm}}

\title{Object detection and autoencoder-based 6D pose estimation for Highly cluttered Bin picking}

\name{Timon H\"ofer, Faranak Shamsafar, Nuri Benbarka, Andreas Zell \thanks{Financial support for this work is provided by the Federal Ministry for Economic Affairs and Energy of Germany (BMWi) in the project iBinPick (ZF4076504DB8). }}
\address{Cognitive Systems Group, CS Department \\Eberhard Karls Universit\"at T\"ubingen, Germany}

\begin{document}

\maketitle

\begin{abstract}
Bin picking is a core problem in industrial environments and robotics, with its main module as 6D pose estimation. However, industrial depth sensors have a lack of accuracy when it comes to small objects. Therefore, we propose a framework for pose estimation in highly cluttered scenes with small objects, which mainly relies on RGB data and makes use of depth information only for pose refinement. In this work, we compare synthetic data generation approaches for object detection and pose estimation and introduce a pose filtering algorithm that determines the most accurate estimated poses. We will make our real dataset for object detection available with the paper.  

\end{abstract}

\begin{figure}[t]
	\begin{subfigure}{.49\linewidth}
		\centering
		\includegraphics[width=0.95\linewidth]{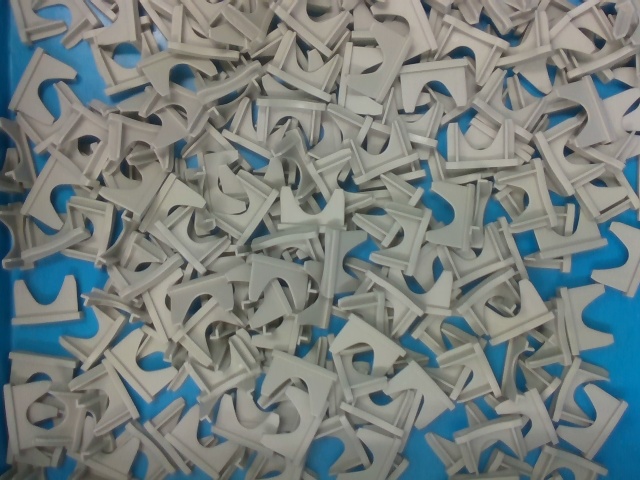}
		\caption{}
		\label{fig:real}
	\end{subfigure}
	\begin{subfigure}{.49\linewidth}
		\centering
		\includegraphics[width=0.95\linewidth]{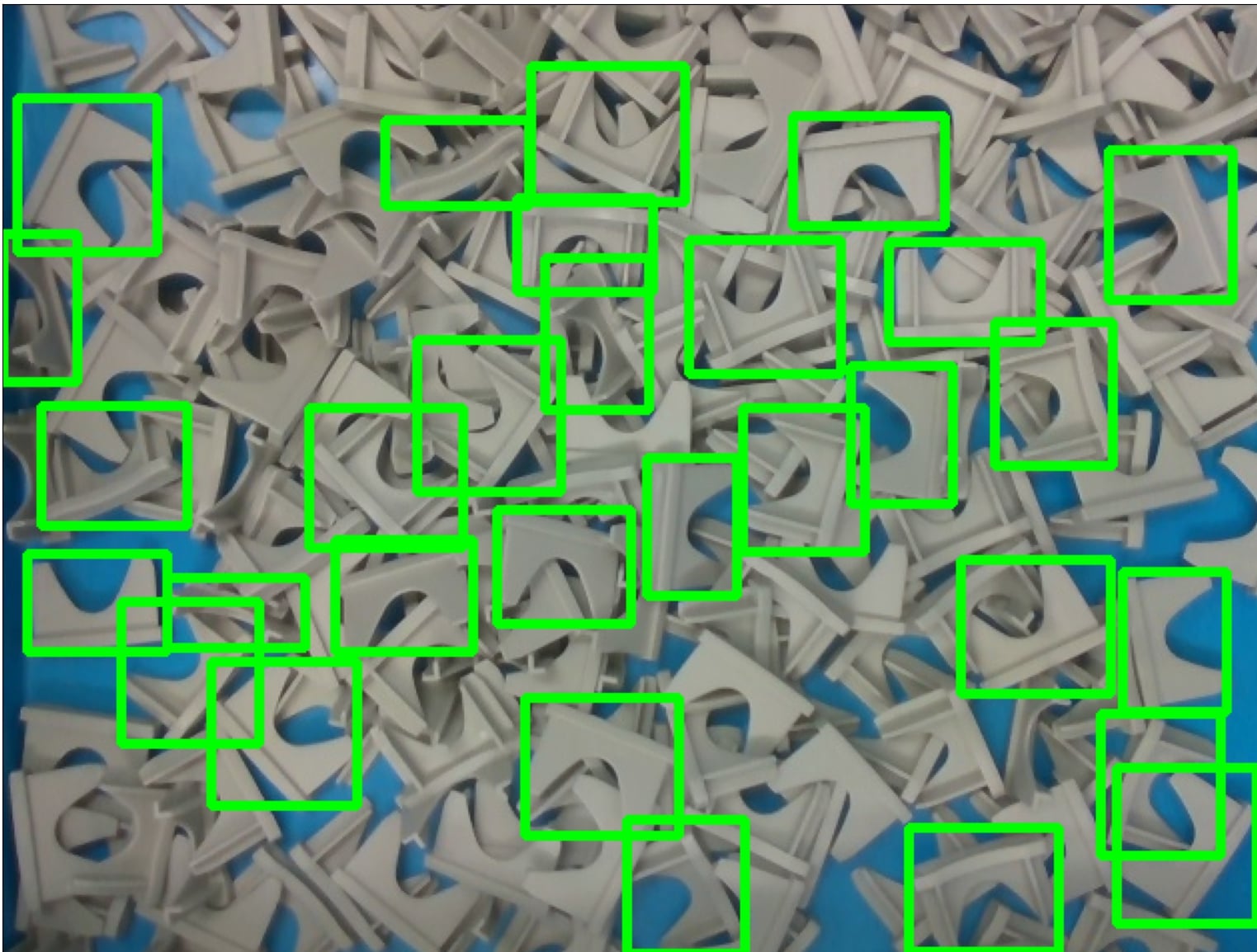}
		\caption{}
		\label{fig:det}
	\end{subfigure}
	\begin{subfigure}{.49\linewidth}
		\centering
		\includegraphics[width=0.95\linewidth]{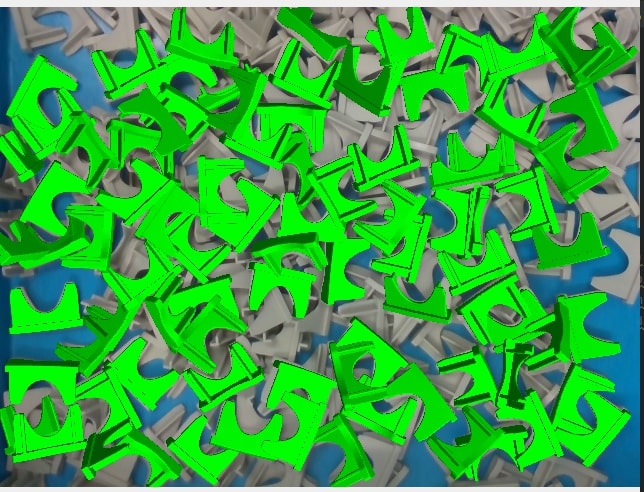}
		\caption{}
		\label{fig:pose_all}
	\end{subfigure}
	\begin{subfigure}{.49\linewidth}
		\centering
		\includegraphics[width=0.95\linewidth]{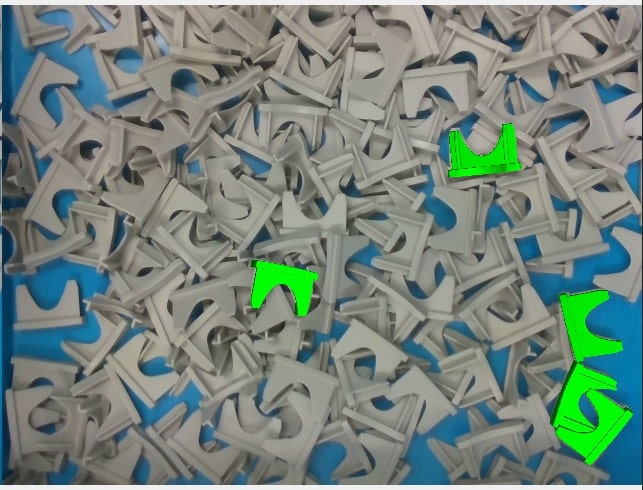}
		\caption{}
		\label{fig:pose_5}
	\end{subfigure}
	\caption{(a) A sample image of a cluttered bin captured by a Microsoft Azure Kinect RGB-D camera. (b) Detection results, limited to 30 objects, to be visually recognizable. (c) Pose estimation results for all the detected objects. (d) The best five selected poses based on the filtering algorithm.}
	\label{fig:pipeline}\vspace{-0.2cm}\vspace{-0.2cm}
\end{figure}
\section{Introduction}
\label{sec:intro}
Bin picking is a major automation task with various applications in industrial sectors. The core starting problem of this work is the 6D pose estimation of instances.
To tackle this problem, an RGB-D or depth camera is usually installed on top of the bin. There are existing solutions to bin picking of large objects, mostly using local invariant features \cite{abbeloos2016point,liu2018point} or template-matching algorithms \cite{hinterstoisser2011multimodal}, which rely on the computationally expensive evaluation of many pose hypotheses. Moreover, local features do not perform well for texture-less objects, and thus, template-matching often fails in heavily cluttered scenes with severely occluded objects. Additionally, depth sensors are often more sensitive to lighting variations than RGB cameras \cite{sundermeyer2018implicit}. Most importantly, for small objects, the depth information is often insufficient to get accurate pose estimates. Therefore, in this work, we focus on RGB-based convolutional neural networks, which make use of depth information only for pose refinement.\\
\indent Accordingly, one of the important issues for training a deep network is labeling the training dataset, which requires high effort for tasks like 6D pose estimation \cite{hodan2017t}. Given a CAD model of the object, which is usually available in the industry, generating a synthetic dataset is possible. However, training on only synthetic 2D images of the CAD models does not generalize well to real data. Hence, more insightful techniques are required to bridge the gap between simulation and reality \cite{sundermeyer2018implicit}.\\
\indent Generally, a state-of-the-art object detector is first used to recognize individual objects, and the resultant cropped images are passed to the pose estimator.
Following \cite{labbe2020cosypose,joffe2019pose},
we use Mask R-CNN \cite{he2017mask} for object detection. As for the task of pose estimation, we consider an augmented autoencoder \cite{sundermeyer2018implicit}, since it has demonstrated good performance in bin picking of deformable products \cite{joffe2019pose}. Sample results of our proposed method are displayed in Fig. \ref{fig:pipeline}. 
The main contributions of this work are as follows:
\begin{enumerate}
	\item We present a comprehensive framework from creating a synthetic dataset to the prediction of the 6D pose estimates in bin picking scenarios, where no real labeling is needed.
	\vspace{-0.22cm}
	\item We show that a more realistic renderer for data generation significantly improves the performance on heavily cluttered piles.
	\vspace{-0.225cm}
	\item We present a pose filtering scheme to select the best pose predictions.
	\vspace{-0.23cm}
	\item We give an analysis of how the performance of the autoencoder can be improved in bin picking scenarios.

\end{enumerate}
The remainder of the paper is as follows: after presenting the related work in section 2, we explain our methodology in section 3. Experimental results are discussed in section 4 and we conclude the paper in section 5.

\section{RELATED WORK}
\label{sec:related}

The object pose estimation problem in bin picking scenarios has been investigated using local invariant features (e.g., point pair features \cite{abbeloos2016point,liu2018point}) and template-matching \cite{hinterstoisser2011multimodal}. However, these approaches do not show acceptable performance in bin picking from a cluttered pile of textureless small objects. Recently, convolutional neural networks \cite{sundermeyer2018implicit,labbe2020cosypose,park2019pix2pose,li2019cdpn} have proven to be promising in the BOP2020 challenge \cite{BOP2020} on 6D pose estimation, even surpassing the depth-based methods. They also tend to be faster, mostly with a runtime of less than one second. These methods mainly depend on an object detection phase, which is achieved using a state-of-the-art object detector (Mask R-CNN \cite{he2017mask}, RetinaNet \cite{lin2017focal}, Faster R-CNN \cite{ren2016faster}).\\
\indent The most important requirements for these deep networks are labeled datasets. 
As the effort of labeling 6D poses in cluttered scenes is high and demands a complex setup \cite{hodan2017t},
some works \cite{kehl2017ssd,sundermeyer2018implicit} have proposed training on synthetic images rendered from a 3D model. To bridge the gap to reality, random augmentations and domain randomization techniques have been applied \cite{pashevich2019learning,tobin2017domain}.
As a different solution to this problem, a more realistic data generation using a physics engine has been proposed in \cite{denninger2019blenderproc}.
In our work, we benefit from this approach to generate photorealistic cluttered piles and propose the full framework for object detection and pose estimation.\\
\indent Moreover, once the general pipeline is given, the predicted poses can be further refined using a pose refinement method. Previously, this step has been achieved \cite{labbe2020cosypose,sundermeyer2018implicit} by the ICP algorithm \cite{zhang1994iterative}. As the ICP-based methods show slow performance, we show that incorporating the depth information into the pose estimation procedure, achieves comparable results. Besides, a filtering algorithm 
is applied to choose the best poses among the estimated ones.

\begin{figure}[t]
	\begin{center}
	\begin{subfigure}{.32\linewidth}
		\centering
		\includegraphics[width=0.95\linewidth]{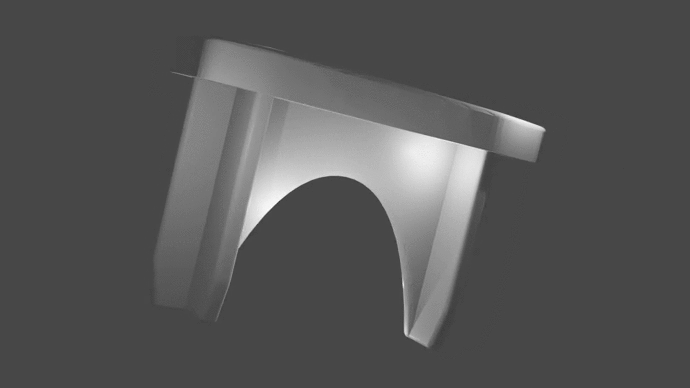}
		\caption{Object 1}
		\label{fig:obj2}
	\end{subfigure}
	\begin{subfigure}{.32\linewidth}
		\centering
		\includegraphics[width=0.95\linewidth]{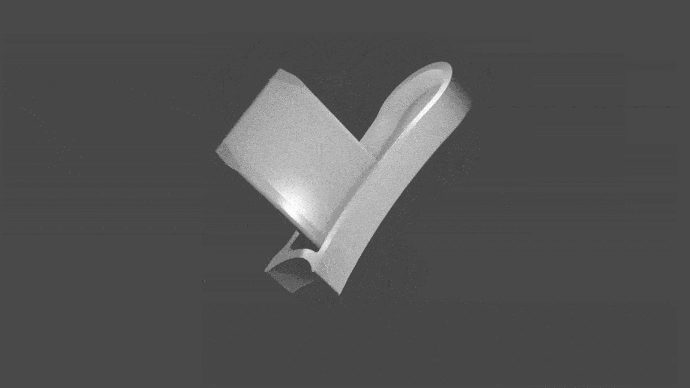}
		\caption{Object 2}
		\label{fig:obj3}
	\end{subfigure}
	\caption{The objects of interest are grey plastic pieces with sizes of $2.3\times3.6\times0.8 \text{cm}^3$ and $1.5\times 2.7\times 0.9\text{cm}^3$, respectively.}
	\label{fig:objects}
\end{center}\squeezeup
\end{figure}

\section{Methodology}
\label{sec:pagestyle} 
In this work, we consider heavily cluttered and occluded scenes of small industrial objects. Here, we explain the methods for dataset generation, followed by the full framework for object detection and pose estimation. \vspace{-0.2cm}
\subsection{Dataset generation}
Creating a synthetic dataset can be achieved by using the CAD models of the objects. Since our pipeline has two main tasks, we need to create a dataset for both, object detection and pose estimation.
\vspace{-0.02cm}
\vspace{-0.3cm}

\subsubsection{Dataset for object detection} 
\begin{figure}[t]
	
	\begin{subfigure}{.49\linewidth}
		\centering
		\includegraphics[width=0.95\linewidth]{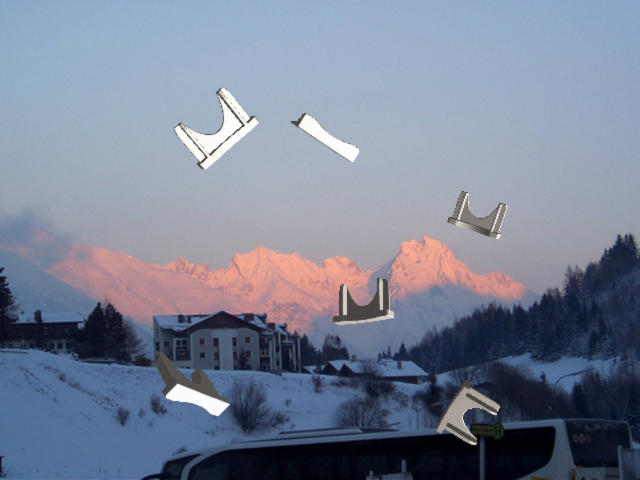}
		\caption{Naive dataset}
		\label{fig:dataset1}
	\end{subfigure}
	\begin{subfigure}{.49\linewidth}
		\centering
		\includegraphics[width=0.95\linewidth]{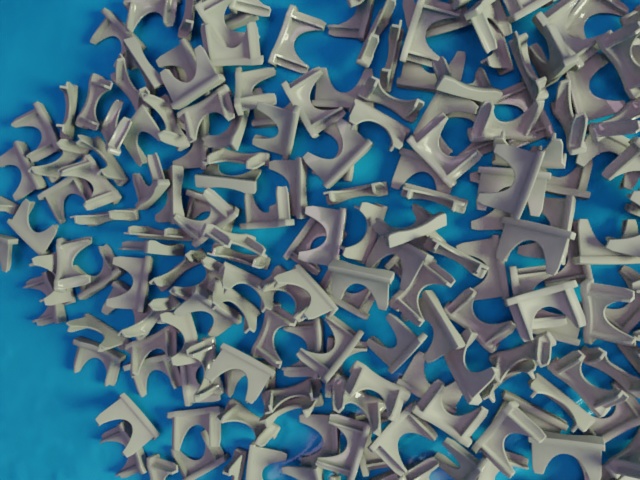}
		\caption{Realistic dataset}
		\label{fig:dataset2}
	\end{subfigure}
	\caption{The synthetic datasets generated with two different pipelines for object 1.}
	\label{fig:datasets}\vspace{-0.25cm}
\end{figure}

As the first approach, we make use of the pipeline in \cite{sundermeyer2018implicit} to generate synthetic images to train the object detector. In particular, a CAD model is used to render the object on a black background. Then, random images from the Pascal VOC dataset \cite{Everingham15} are added as a background, followed by random augmentations strategies. We create 60K images per object with 5-20 instances per scene. Due to their simplicity, we call these images \quotes{naive dataset}. \\
\indent For the second approach, we employ BlenderProc \cite{denninger2019blenderproc} to generate more realistic synthetic images. 
BlenderProc utilizes a physics engine to make the synthetic data look more realistic. Furthermore, it uses different lighting effects, object materials and applies physics and collision checking as well. 
With BlenderProc, we generate for each object type 20K images with 30 instances and 5K images with 300 objects. The camera configuration is sampled in a range of $20^\circ$ around the top of the scene with a height between 27-33cm.
We call this the \quotes{realistic dataset}.\\
\indent In this work, we consider two industrial objects (see Fig. \ref{fig:objects}).
Sample images of the naive and realistic images for object 1 are depicted in Fig. \ref{fig:datasets}.
 \vspace{-0.3cm}
\subsubsection{Dataset for pose estimation}
Similar to the naive dataset in the previous section, we generate images with corresponding 6D pose annotations, with an additional step of image cropping around the objects. We also compare the original pipeline results against a new dataset, where we render multiple objects in the image crops. In Fig. \ref{fig:2coe}, samples of different training data are displayed on the left side. While the top image shows one single object in each image crop, the bottom one includes multiple objects.
 \vspace{-0.3cm}
\subsubsection{Test dataset for object detection}
\begin{figure}[t]
	\begin{center}
	\begin{subfigure}{.39\linewidth}
		\centering
		\includegraphics[width=0.95\linewidth]{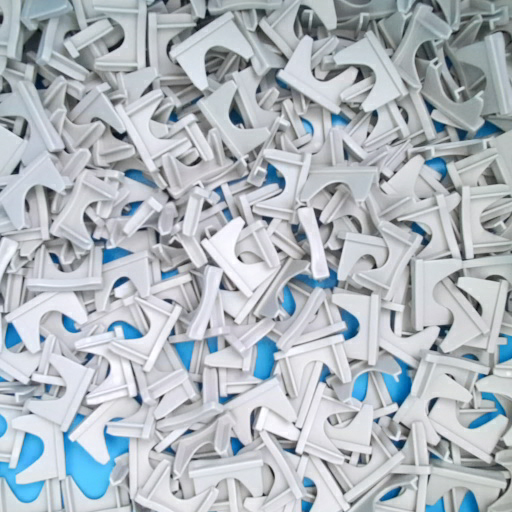}
		\caption{Real test dataset 1}
		\label{fig:testdataset1}
	\end{subfigure}
	\begin{subfigure}{.39\linewidth}
	\centering
	\includegraphics[width=0.95\linewidth]{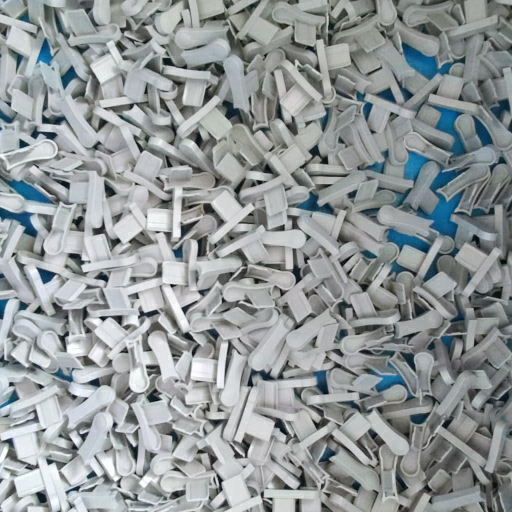}
	\caption{Real test dataset 2}
	\label{fig:testdataset2}
	\end{subfigure}
	\caption{Examples of our labeled test dataset in real scenarios.}
	\label{fig:testdataset}
\end{center}\squeezeup
\end{figure}

To evaluate the object detector, we captured 50 real images per object model with more than 100 instances in the bin. The images were taken using a Microsoft Azure Kinect camera mounted at a height of 30cm over the bin (see Fig. \ref{fig:testdataset}).\label{testdet}
 \vspace{-0.4cm}

\subsubsection{Test dataset for pose estimation}
While we show qualitative results in real-world scenarios, the quantitative results are reported on a synthetic dataset, because only for this we have full ground truth data. The autoencoder is trained on synthetic data following the pipeline in \cite{sundermeyer2018implicit}, and we create the test dataset with BlenderProc \cite{denninger2019blenderproc}.To be more precise, BlenderProc generates 3D scenes, whereas the pipeline in \cite{sundermeyer2018implicit} creates augmented 2D images. Since the data distributions of these synthetic datasets are different, we will show that training the pose estimator on the naive dataset and testing on the photorealistic images leads to suitable performance.\\ 
\indent As such, for each object, we created one dataset consisting of 1K images with 300 objects. The camera is located at 30cm on top of the bin ground. We choose a blue background to make it visually comparable to our real settings. 

\subsection{Object detection}
In general, any state-of-the-art object detector can be used (Faster R-CNN \cite{ren2016faster}, RetinaNet \cite{lin2017focal}, SSD \cite{liu2016ssd}) for object detection. However, these methods only predict the bounding boxes. Therefore, we choose Mask R-CNN, which has the advantage of predicting the segmentation masks of the object as well. This can be used for pose refinement \cite{wong2017segicp} by segmenting the point clouds of the objects. In addition, we can compare the pose estimation results when, instead of the whole bounding box, only the pixels visible in the segmentation mask are given to the pose estimation module.
Given an image, we predict a set $\mathcal{D}$ of object detections.

\subsection{Pose estimation}
\begin{figure}
\centering
\includegraphics[width=0.95\linewidth]{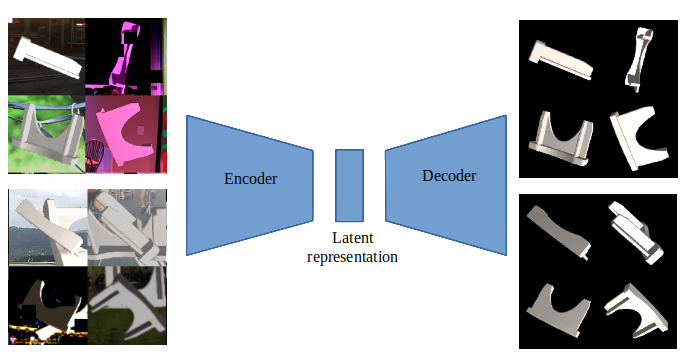}
\caption{The architecture of the autoencoder. The autoencoder is trained to map the augmented images to the original image. On the left, there are the two different types of training data.}
\label{fig:2coe}\vspace{-0.2cm}
\end{figure}

To receive pose estimates from the set $\mathcal{D}$ of the detections, we resort to the autoencoder network presented in \cite{sundermeyer2018implicit}. As the autoencoder's training procedure is based on only synthetic data, it is more applicable to new industrial settings, where no labeled data exists. In addition, the autoencoder has demonstrated good performance in pick-and-place tasks \cite{deng2020self, joffe2019pose}.
Its method of operation is as follows:
the autoencoder is a dimensionality reduction technique trained to extract a 3D object from image crops (see Fig. \ref{fig:2coe}).
After training, we create a codebook to determine the rotation of the object. The codebook is the set of all latent representations of the discretized 3D rotations that cover the whole SO(3).\\
\indent At test time, the image crops from the set $\mathcal{D}$ are fed into the encoder. The resulting latent representation $z_{test}$ is then compared with all the latent representations $(z_i)$ from the codebook via a k-NN search, with the similarity function as:
\begin{align}
	cos_i = \frac{z_iz_{\text{test}}}{||z_i||\, ||z_{\text{test}}||}.\label{gl:cosine}
\end{align}
We then choose the rotations with the highest cosine similarities.
\subsection{Selecting the best pose estimates}
 In cluttered scenes, we have pose estimations of several hundred objects. These will be used for the picking task, and as the robot will only pick one object at a time, we are only interested in the k-top pose estimates, and the question arises, how to select the best k-pose estimates. While one can sort the pose estimates regarding the scores given by Mask R-CNN, or by the highest cosine similarities \eqref{gl:cosine},
we define a new selecting method that compares the depth of the original image with the depth of the rendered image.
Let $A^i_1$ be the predicted segmentation mask of object $i$.
Given a predicted 6D pose ($\bar{t},\bar{R})$, we render the depth image $\bar{\Omega}$ and we define the set of pixel $A^i_2:= \{(p,q): \; |\Omega(p,q)-\bar{\Omega}(p,q)|<m\}$, for some margin $m$ and where $\Omega$ represents the real depth image. $A^i_3$ is the segmentation mask of the rendered image. With these definitions, we can build the intersection:
	$\mathcal{A}^i:=A^i_1 \cap A^i_2 \cap A^i_3$.
For each pose estimate we calculate the depth error as follows:
\begin{align}\label{e_i}
e_i= \sum_{(p,q)\in \mathcal{A}_i}|\Omega(p,q)-\bar{\Omega}(p,q)|
\end{align}
In the next section, we compare the different approaches.
\section{Experimental results}

\subsection{Experiments on object detection}
 We fine-tuned a pretrained Mask R-CNN with a ResNet-50 backbone for 15 epochs with an initial learning rate of 0.001 and a mini-batch size of 4 images. The learning rate was reduced by a factor of 10 at epochs 3, 6, 9 and 12. Stochastic gradient descent (SGD) with momentum (0.9) and weight decay (0.0005) was used for optimization. Our work is based on the torchvision implementation of Mask R-CNN \cite{marcel2010torchvision}. In Table. \ref{tab:detection}, the accuracies of object detection in terms of $AP_{50}$ (the average precision with IoU thresholded at 0.50), $AP_{50:95}$ and $AR^{max=100}$ (the average recall with 100 detections per image) are tabulated. While training on the naive dataset does not generalize well to our heavily cluttered real scenarios, Mask R-CNN trained on the realistic dataset considerably boosts the performance.

\begin{table}[t]
	\centering
	\begin{tabular}{c||c|c|c|c}
		Object &  dataset & \footnotesize 	$\textrm{AP}_{50:95(\%)}$& $\footnotesize 	\textrm{AP}_{50(\%)}$ & \footnotesize 	$\textrm{AR}^{\text{\tiny max}=100(\%)}$ \\
		\hline
		\multirow{2}*{Object 1} & naive& 9.3&  12.5 & 9.6   \\	\cline{3-5}	
		& realistic& \textbf{66.8} & \textbf{82.8} & \textbf{80.3}   \\ \hdashline

		\multirow{2}*{Object 2} & naive & 0.9 &1.0   & 0.1  \\	\cline{3-5}	
		& realistic & \textbf{50.4} & \textbf{68.7}& \textbf{59.2}    \\

		\hline
	\end{tabular}
	\caption{Object detection results after training Mask R-CNN on different synthetic datasets}
	\label{tab:detection}\vspace{-0.2cm}
\end{table}
\subsection{Experiments on pose estimation}

To this goal, we trained the autoencoder with a latent space size of 128. We chose the L2 loss function, a learning rate of 0.0001 and used the Adam optimizer with a batch size of 32 and trained it for 40K iterations. The pose error metrics used for evaluation are the Visible Surface Discrepancy (VSD), the Maximum Symmetry-aware Surface Distance (MSSD) and the Maximum Symmetry-aware Projection Distance (MSPD), that are being used in the BOP2020 challenge \cite{BOP2020}. An estimated pose is considered as correct w.r.t. the pose-error function $e$ if $e<\theta_e$, where $e\in \{e_{\text{VSD}},e_{\text{MSSD}},e_{\text{MSPD}}\}$ and $\theta_e$ is the threshold of correctness. We used the same values for $\theta_e$ as in \cite{BOP2020} to calculate the average recall rates $AR_{\text{VSD}}$
, $AR_{\text{MSSD}}$ and $AR_{\text{MSPD}}$. The performance of the method on a dataset is measured by the Average Recall $AR=\big(AR_{\text{VSD}} +AR_{\text{MSSD}}+ AR_{\text{MSPD}} \big)/3$. \\
\indent In Table \ref{tab:sorting}, we compare the results of the different methods on selecting the best five pose estimates. In these experiments, we did not make use of ICP, but we took the depth measurement at the center of the object. It shows that sorting according to the vector $(e_i)$ in \eqref{e_i}, results in superior performance compared to other approaches.\\
\indent The experiments in Table \ref{tab:refinment} are conducted using the selection method defined by the vector $(e_i)$. We compare the results, when testing with only RGB information, using the depth measurement at the object center and the improvement through ICP refinement. While ICP refinement increases the performance slightly, the refinement of several hundred poses per image takes time, making it impractical for the usage in real-time settings. The experiments to reduce the noise in cluttered scenes, like feeding only the pixels visible in the mask to the autoencoder, or training the autoencoder with multiple objects, have shown, against our intuition, a worse performance than the normal pipeline. Note how incorporating the depth has improved the accuracy. We have shown qualitative results in the supplementary materials.

\begin{table}[t]
	\centering
	\begin{tabular}{c||c|c|c}
		\multirow{2}*{Object} & sorted by Mask& sorted by max & sorted by depth \\
		&R-CNN scores&cosine sim.& differences $(e_i)$\\
		\hline
		Object 1 & 0.509 & 0.394 &\textbf{0.812}  \\	

		\hdashline

		Object 2 & 0.533 & 0.449  & \textbf{0.633}  \\

		\hline
	\end{tabular}
	\caption{Average recall of top 5 pose estimates sorted by three different approaches for selecting the best estimates}
	\label{tab:sorting}\vspace{-0.3cm}
\end{table}

\begin{table}[t]
	\centering
	\begin{tabular}{c||c|c|c cc}
		\multirow{2}*{Object} &RGB&  \multicolumn{4}{c}{RGB + depth} \\
		&&+ ICP& normal& mult.-obj.& mask\\
		\hline
		Object 1 &0.691 &\textbf{0.829}  &0.812&0.790  &0.798\\	
		\hdashline
		Object 2 & 0.348  & \textbf{0.703} & 0.633& 0.627 &0.614 \\
		\hline
		time (s) &\textbf{0.699}  & 18.39 & &\textbf{0.697 } \\
		\hline
	\end{tabular}
	\caption{Average recall of top 5 pose estimates using ICP and depth measurements and combinations.}
	\label{tab:refinment} \vspace{-0.3cm}
\end{table}
\section{Conclusion}
In this paper, we investigated the task of bin picking from piles of crowded, small-sized and identical objects. In particular, we explored the main required vision modules for this challenging problem, i.e. object detection and pose estimation. For each task, we employed convolutional neural networks, which are trained on two types of generated synthetic datasets. Experimental results on synthetic and real images show that the proposed comprehensive framework, from dataset generation to pose estimation, is promising for industrial bin picking.
\bibliographystyle{IEEEbib}
\bibliography{Literaturzeugs_new}

\end{document}